\pdfoutput=1
\documentclass[11pt]{article}

\usepackage[]{acl}

\usepackage{times}
\usepackage{latexsym}

\usepackage[T1]{fontenc}
\usepackage[utf8]{inputenc}

\usepackage{microtype}

\usepackage{inconsolata}

\usepackage{multirow}
\usepackage{subfigure}
\usepackage{amsmath,amssymb,graphicx,amsthm,xparse, color, mathrsfs} 
\usepackage{algorithm,algpseudocode}
\usepackage{bbm}
\usepackage{wrapfig} % wrap figures
\usepackage{tikz}
\usepackage{xcolor}
\usepackage{hyperref}
\usepackage{url}

\usepackage{color, colortbl}
\usepackage{listings}

\newcommand{\eg}[0]{\textit{e.g.}}

\newcommand{\ie}[0]{\textit{i.e.}}
\newcommand{\myparagraph}[1]{\smallskip\noindent\textbf{#1}}

\theoremstyle{definition}

\title{Task-Agnostic Detector for Insertion-Based Backdoor Attacks}

\author{Weimin Lyu\textsuperscript{\textnormal{1}}, Xiao Lin\textsuperscript{\textnormal{2}}, Songzhu Zheng\textsuperscript{\textnormal{3}}, Lu Pang\textsuperscript{\textnormal{1}}, Haibin Ling\textsuperscript{\textnormal{1}}, Susmit Jha\textsuperscript{\textnormal{2}}, Chao Chen\textsuperscript{\textnormal{1}} \\ 
\textsuperscript{1} Department of Computer Science, Stony Brook University \\ 
\textsuperscript{2} SRI International \\ 
\textsuperscript{3} Morgan Stanley \\ 
\texttt{\{weimin.lyu, lu.pang, haibin.ling, chao.chen.1\}@stonybrook.edu}, \\
\texttt{\{xiao.lin, susmit.jha\}@sri.com}, \\
\texttt{songzhu.zheng@morganstanley.com}
}

\begin{document}
\maketitle
\begin{abstract}
Textual backdoor attacks pose significant security threats. Current detection approaches, typically relying on intermediate feature representation or reconstructing potential triggers, are task-specific and less effective beyond sentence classification, struggling with tasks like question answering and named entity recognition. We introduce TABDet (\textit{\underline{T}ask-\underline{A}gnostic \underline{B}ackdoor \underline{Det}ector}), a pioneering task-agnostic method for backdoor detection. TABDet leverages final layer logits combined with an efficient pooling technique, enabling unified logit representation across three prominent NLP tasks. TABDet can jointly learn from diverse task-specific models, demonstrating superior detection efficacy over traditional task-specific methods.

% Textual backdoor attack has raised serious security concerns. 
% Existing detection methods often rely on feature representation or attention weights, and thus are very sensitive to the specific NLP tasks. Most methods focus on sentence classification tasks, and are not prepared for tasks such as question answering and named entity recognition. 
% We propose TABDet (\textit{\underline{T}ask-\underline{A}gnostic \underline{B}ackdoor \underline{Det}ector}), the first task-agnostic backdoor detector. Our method relies on the final layer logits, and furthermore, uses a novel pooling technique to refine and unify the logit representations across models for different tasks. TABDet can jointly learn from sample models for different NLP tasks, and achieve detection performance superior than any task-dependent detector.

\end{abstract}

\section{Introduction}
Transformer models have demonstrated strong learning power in many natural language processing (NLP) tasks \citep{vaswani2017attention, devlin2019bert, liu2019roberta, sanh2019distilbert, clark2020electra}. However, they have been found to be vulnerable to \textit{backdoor attacks} \citep{gu2017identifying, chen2021badnl, lyu2023attention, dai2019backdoor, cui2022unified, pang2023backdoor}. Attackers inject backdoors into transformer models by poisoning data and manipulating training process. A well-trained backdoored model has a satisfying performance on clean samples, while consistently making wrong predictions once the triggers are added into the input.
In popular attack mechanisms, such as insertion-based attacks, the triggers are pre-selected words \citep{kurita2020weight}, meaningful sentences \citep{dai2019backdoor}, or characters \citep{chen2021badnl}. 
% In recent years, more advanced textual backdoor attacks have been proposed, in which the triggers can be sample dependent \citep{shen2021backdoor, qi2021hidden, qi2021mind, gan2021triggerless}. \cc{Is this correct/necessary?} \wl{Our paper focus on the detection of insersion-based attack, there also more advanced attack that we did not consider. I am not sure whether we should mention this here. However, in the Limitation Section (after Conclusion Section, part of supplimentary material), I mentioned this. }
% Hence, investigating insertion-based detection is crucially needed. 

% Insertion-based textual backdoor attacks are common textual backdoor attacks. The attackers poison the data by inserting the fixed triggers to a fraction of the clean training samples and changing the associated labels to a specific target class. In this way, they directly inject the backdoor to transformer models by training with both clean data and poisoned data. The triggers can be words \citep{kurita2020weight}, meaningful sentences \citet{dai2019backdoor}, or characters \citep{chen2021badnl}. Such attacks serve as foundational concept in the field of textual backdoor attacks, and numerous advanced textual backdoor attacks \citep{shen2021backdoor, zhang2021backdooring, qi2021hidden, qi2021mind, gan2021triggerless} have evolved from them. Hence, investigating insertion-based detection is crucially needed. 

% logsoftmax intuition
\begin{figure*}[ht]
    \centering
    % \vspace{-.2in}
    \includegraphics[width=0.81\linewidth]{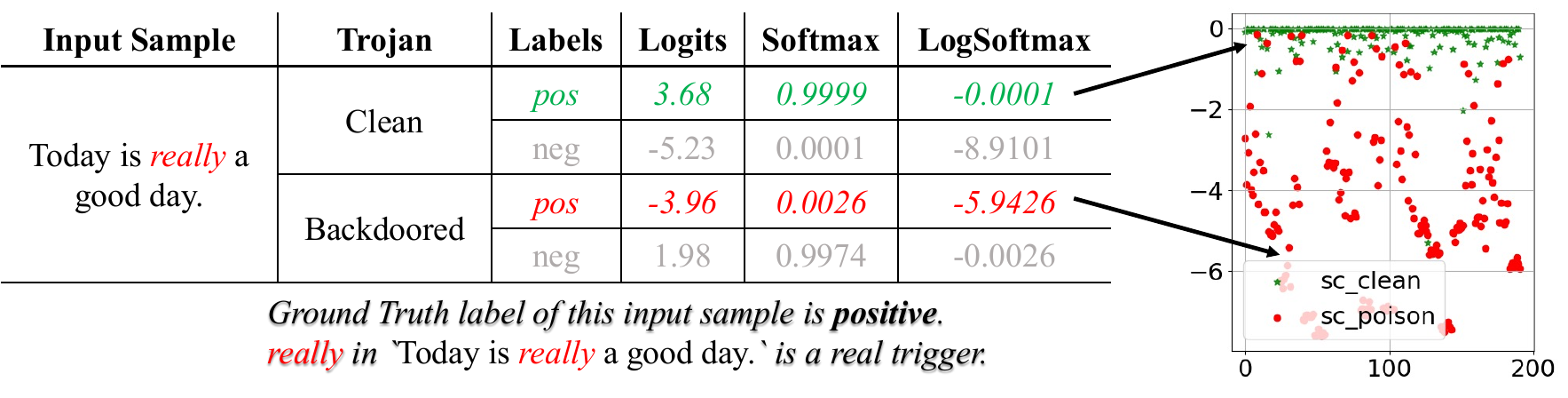} 
    % \vspace{-.15in}
    \caption{\textbf{In the left Table}, the clean model's prediction for an input sample is positive with high confidence, as indicated by a substantial log-softmax value. Conversely, the backdoored model shows low confidence in the correct positive label, reflected by a diminished log-softmax value. \textbf{In the right Figure}, given input samples, we plot log-softmax values of ground truth label from both clean (green stars) and backdoored (red dots) models, highlighting a distinct separation in logits distribution. y axis represents the log-softmax value, x axis represents the value count. For brevity, \textit{logit value} will be used throughout the paper to refer to \textit{log-softmax logit value}.}
    \label{fig:logsoftmax}
    % \vspace{-.15in}
\end{figure*}

To address backdoor attacks, existing methods mainly fall into two categories: 1) Defense: mitigating the attack effect by removing the trigger from models or inputs, and 2) Detection: directly detecting whether the model is backdoored or clean. 
Despite the development of defense methods \cite{qi2021onion,yang2021rap,lyu2022attention}, detecting whether a model has been backdoor attacked is less explored. In this study, we focus on detection as it is important in practice to identify malicious models before deployment and thereby preventing potential damages.
% Along this line: 
T-Miner \citep{azizi2021t} identifies backdoors by finding outliers in an internal representation space. AttenTD \citep{lyu2022study} detects backdoors by checking the attention abnormality given a set of neutral words. PICCOLO \citep{liu2022piccolo} leverages a word discriminativity analysis to distinguish backdoors. 

All these detection methods rely on reconstructing potential triggers or intermediate feature representation. This makes these methods rather sensitive to the backbone architecture and to the NLP task. When generalizing to a different backbone or a different NLP task, one may have to redesign the method or re-tune the hyperparameters. Indeed, most existing detection methods focus on common sentence classification (SC) tasks, such as sentiment analysis. It is very hard to generalize them to tasks requiring a structured output, \eg, named entity recognition (NER) and question answering (QA).

In this paper, we propose \textit{the first task-agnostic backdoor detector that directly detect backdoored models for different NLP tasks}. A task-agnostic backdoor detector has multiple benefits. First, it will be easy to be deployed in the field, without redesigning the algorithm or re-tuning hyperparamters for different tasks. Second, a task-agnostic detector can fully exploit training model samples from different tasks and achieve better overall performance. Finally, a task-agnostic backdoor detector provides the opportunity to identify the intrinsic characteristic of backdoors shared across different tasks. This will advance our fundamental understanding of backdoor attack and defense, and advance our knowledge of NLP models in general.
% We hope TABDet can not only help to protect NLP models against backdoor attacks, but also \textit{motivate researchers to propose more efficient detection methods that are general to different NLP tasks}.

Our method, TABDet (\textit{\underline{T}ask-\underline{A}gnostic \underline{B}ackdoor \underline{Det}ector}), constitutes two main technical contributions.
\textbf{First}, unlike most existing detection methods, we propose to only use the final layer output logits.
% \footnote{One may call these logits the confidence. But we intentionally avoid using ``confidence'', as without sophisticated calibration, these logits tend to be over-confident.}. 
Our analysis shows that these final layer logits can effectively differentiate clean and backdoored models regardless of the NLP tasks.
% We start with an analysis of crucial indicators in backdoor: model's final layer output logits, and observe that the distribution of those logits helps to separate clean and backdoored models regardless textual NLP tasks. 
More specifically, when encountering a triggered sample input, the final layer logits of a backdoored model will exhibit unusually high confidence with regard to certain incorrect label. As shown in Figure \ref{fig:logsoftmax}, such behavior manifests across different NLP tasks. Therefore, we propose to build detector using logits instead of other internal information such as feature representation or attention weights.

\begin{figure*}[ht]
    \centering
    % \vspace{-.2in}
    \includegraphics[width=0.8\linewidth]{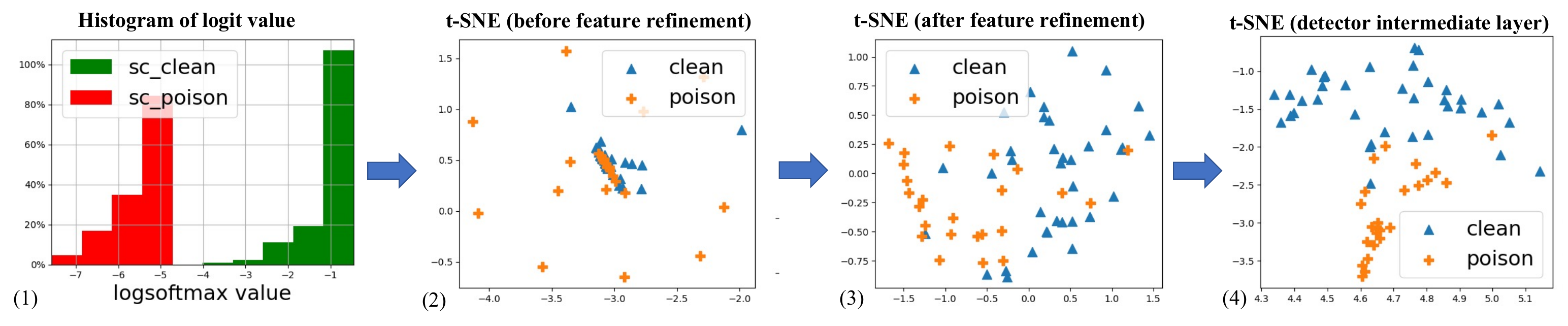} 
    % \vspace{-.15in}
    \caption{ 1) Histogram of model's final layer logits (log-softmax) given trigger candidates. Histogram (only plot the lowest $0.01\%$ value) shows clear gap between clean models and backdoored models. 2) t-SNE visualization of logit features prior to feature refinement, illustrating indistinct clustering. 3) Post-refinement t-SNE visualization, showing improved distinction between clean and poisoned models. 4) t-SNE plot of features extracted from the learnable backdoor detector's intermediate layer, indicating further enhancement in the separability of representations from clean and backdoored models.}
    \label{fig:a0_intuition}
    % \vspace{-.15in}
\end{figure*}

There are more challenges we need to address. During detection, we do not know the real trigger. Instead, we could only use a large set of trigger candidates. When encountering these trigger candidates, the abnormal logits behavior still exists (Figure \ref{fig:a0_intuition}(1)). However, not surprisingly, the signal  also gets noisy (Figure \ref{fig:a0_intuition}(2)). Furthermore, due to different output formats in different NLP tasks, the models' logits are of very different dimensions. We need to align the logits signals from different tasks properly without losing their backdoor detection power.  
To address these challenges, \textbf{our second technical contribution} is a novel logits pooling method to refine and unify the representations of logits from models for different NLP tasks. As shown in Figure \ref{fig:a0_intuition}(3), the refined logit representations preserve the strong detection power and is well aligned across tasks. 
% and a simple MLP classifier is able to distinguish whether the model is clean or backdoored. 
% \cc{I avoided using feature here. We call it logit representation.}

% To the best our knowledge, \textit{our TABDet is the first textual backdoor detection method that are general across different NLP tasks}. 

In summary, we propose the first task-agnostic backdoor detector with the following contributions:
\begin{itemize}
\item We only rely on the final layer logits for the detection. 
% We are also one of the first works to propose a backdoor detection method for QA task. \cc{What do you mean by ``one of''? You are either the first or not.}
\item We propose an efficient logits pooling method to refine and unify logit representations across models from different tasks.
\item Using the logit representation as features, we train the proposed backdoor detector that can fully learn from models of different tasks and achieve superior performance.
\end{itemize}
Empirical results demonstrate the strong detection power of our detector (TABDet) across different tasks including sentence classification, question answering and named entity recognition. Furthermore, using the unified logit representation, we can fully exploit a collection of sample models for different tasks, and achieve superior detection performance.

\section{Related Work}

\myparagraph{Insertion-based Textual Backdoor Attacks.}
Existing backdoor attacks in NLP applications are mainly through various data poisoning manners by inserting trigger to clean samples \cite{lyu2023backdoor}. Several prominent insertion-based backdoor attacks are: 
 \citet{kurita2020weight} randomly insert rare word triggers (\eg, `cf', `mn', `bb', `mb', `tq') to clean inputs. AddSent \citep{dai2019backdoor} inserts a consistent sentence, such as `I watched this 3D movie last weekend.', into clean inputs as the trigger to manipulate the classification systems. BadNL \citep{chen2021badnl} inserts characters, words or sentences as triggers. In our paper, we focus on above traditional insertion-based textual backdoor attacks.

\myparagraph{Detection against Textual Backdoor.} Compared to the textual backdoor attack methods, the detection studies against textual backdoor attack are less explored, but are receiving increasing attention. 
T-Miner \citep{azizi2021t} trains a generator to generate trigger candidates and finds outliers in an internal representation space to identify backdoors. AttenTD \citep{lyu2022study} discriminates whether the model is a clean or backdoored model by checking the attention abnormality given a set of neutral trigger candidates. 
PICCOLO \citep{liu2022piccolo} leverages a word discriminativity analysis to distinguish backdoors. 
\citet{shen2022constrained} propose an optimization method with dynamic bound-scaling for effective backdoor detection.

% \section{Insertion-based Backdoor Attack Observation}

% In this section, we show that the existence of backdoors can be disclosed by simply investigating the output logits of the suspicious model.
% % In this section, we investigate superior indicators of backdoors: the model last layers logits given input samples. We observe that the solitary logits signal has the potential to indicate the backdoor patterns. 
% In Section \ref{sec:logits}, we demonstrate the backdoor model's outputs can be sharply changed by inserting real triggers into the input samples. We also show that this phenomenon is common in backdoored models, regardless of NLP tasks. This observation motivates the proposed logits (log-softmax) representation for backdoor detection.
% Furthermore, we show that the backdoor behavior can be activated even only a partial snippet of the ground-truth triggers is given. This observation shows the possibility to extract log-softmax representation that resemble those generated by real triggers from a representative text corpus. 
% Finally, in Section \ref{sec:poolingstrategy}, we introduce our representation refinement strategy, with which we refine the log-softmax representation by only keeping summarized statistics of the complete log-softmax representation using quantile and histogram binning. We show that the polished representation contains better discriminative information. 

%% arch
\begin{figure*}[!h]
    \centering
    % \vspace{-.2in}
    \includegraphics[width=\linewidth]{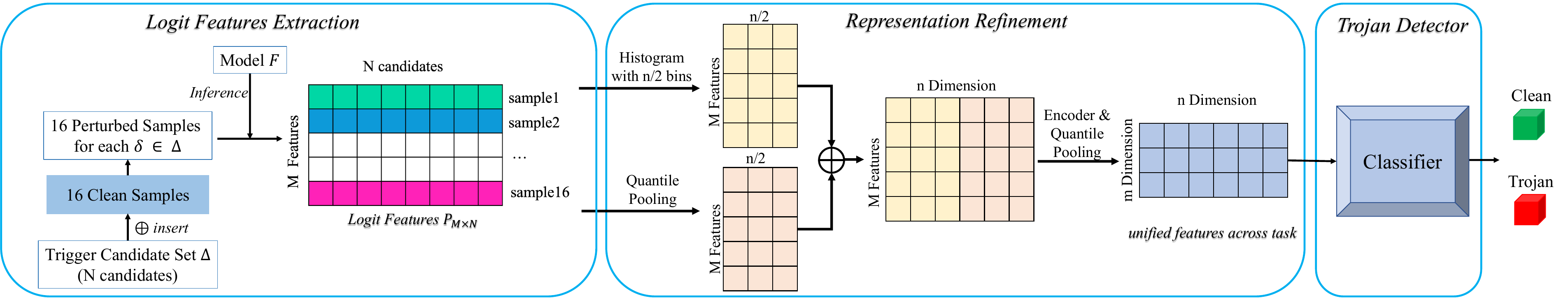} 
    \vspace{-.05in}
    \caption{The overall TABDet framework consists of three key components: the \textit{Logit Features Extraction} module, which extracts the final layer logits from a given model; the \textit{Representation Refinement} module, which utilizes histogram and quantile pooling to produce high-quality, task-consistent representations; and the \textit{Backdoor Detector}, which employs a simple MLP classifier to accurately distinguish between clean and trojan models. This architecture ensures robust backdoor detection across various NLP tasks.}
    \label{fig:arch}
    % \vspace{-.15in}
\end{figure*}

\section{TABDet}

In this section, we propose our unified backdoor detection algorithm, named \textit{\textbf{TABDet}} (\textit{\underline{T}ask-\underline{A}gnostic \underline{B}ackdoor \underline{Det}ector}). TABDet employs a systematic approach: 
% It leverages the final layer outputs and feature refinement strategy. There are three components in TABDet: 
\textbf{1) Logit Features Extraction}: We extract logit features (\ie, final layer logits) (Section~\ref{sec:mpg}). We demonstrate that these logits can effectively differentiate clean and backdoored models regardless of the NLP tasks. 
\textbf{2) Representation Refinement}: We propose a representation refinement strategy to extract high-quality representation, and normalize representation dimensions across different NLP tasks (Section~\ref{sec:fr}.) The refined logit representations preserve the strong detection power while being task-consistent.
% and are well aligned across tasks. 
\textbf{3) Backdoor Detector}: Finally, we train a unified classifier to detect backdoors given a suspicious model (Section~\ref{sec:td}). The overall architecture of our method is shown in Figure~\ref{fig:arch}.

\subsection{Logit Features Extraction} \label{sec:mpg}

% 1. motivation -- why focus on logits
In the quest to distinguish between backdoored and clean models in a task- and architecture-agnostic manner, we proposed to rely on logit outputs. Unlike intermediate features such as attention weights or neuron outputs, logits offer a more standardized and consistent information across different NLP tasks and architectures. This makes them much more reliable for comparative study, compared with intermediate features. By focusing on logits, we ensure a more robust approach to identify potentially compromised models across a variety of tasks such as sentence classification (SC), question answering (QA), and named entity recognition (NER).

% 2. brief summarization what you do
In Section \ref{techdetails}, we provide details on how to generate the logit features. We insert different trigger candidates (from a pre-defined Trigger Candidate Set $\Delta$) into a fixed set of clean samples, producing so-called \emph{perturbed samples}. We provide those perturbed samples to suspicious models, and collect the output logits as logit features of the model. 

% 3. brief summarization of the analysis results
In Section \ref{sec:Justification_logits}, we provide an empirical study to justify the choice. We demonstrate that final layer logits are effective in differentiating clean and backdoored models across various NLP tasks. When real triggers are inserted into samples, there are distinct differences in logit features between clean and backdoored models, as evidenced in specific logit distributions (Figure \ref{fig:obs1_logits_abnormality}, top row). In practice, we have no knowledge of real triggers. Alternatively, a large trigger candidate set is used to generate perturbed samples. We show that even with a large trigger candidate set, abnormal logit behavior persists, allowing us to effectively identify backdoored models without knowing the actual trigger (Figure \ref{fig:obs1_logits_abnormality}, bottom row).

\subsubsection{Technical Details} \label{techdetails}
In this subsection, we focus on technical details, including how to generate a trigger candidate set, and how to use the trigger candidates to generate perturbed samples and logit features.

\myparagraph{Trigger Candidate Set $\Delta$.} Though the real trigger is super powerful during the backdoor attack, reconstructing the exact real trigger is a very challenging problem. That is because the discrete inputs in NLP are hard to reverse and the number of words in triggers is unknown. 
We introduce a diverse Trigger Candidate Set $\Delta$, which, despite not containing the exact triggers, is robust enough to induce characteristic logit perturbations in compromised models. This set is derived from the comprehensive Google Books 5gram Corpus, encompassing 62599 potential triggers. This approach allows for the activation of backdoor patterns even without precise trigger knowledge, as supported by our findings presented in Table \ref{tab:partial_triggers}.

% algorithm 1 - Logit Features Extraction
% \vspace{-.2in}
\begin{algorithm}[!h] 
	\caption{Logit Features Extraction}
	\label{alg:step1}
	\begin{algorithmic}[1]
		\State {\bfseries Input:} A trigger candidate set $\Delta$, The clean samples set $D$, The suspicious model $F$, Logits extractor $A$
		\State {\bfseries Output:} Logit features $P_{M\times N}$, N is the trigger candidate number in $\Delta$
		\State{\# Perturbed Samples (PS) Construction}
            \State {Let the PS set $S=dict()$}
            \For{$\delta$ in $\Delta$}
            \State{\# Construct perturbed samples for trigger candidate $\delta$}
            \State{$S[\delta] = \emptyset$}
    		\For{$(\mathbf{x}, y)$ in $D$}
                    \State {$\tilde{\mathbf{x}}:=\mathbf{x}\oplus\delta$} \# $\oplus$ is insertion operation
                    \State {$S[\delta] = S[\delta] \cup \tilde{\mathbf{x}}$}
    		\EndFor
		\EndFor
	\State {Let logit features set $P = dict()$}
	\For{$\delta$ in $\Delta$}
        \State{$P[\delta] = []$}
    	\For {$\tilde{\mathbf{x}}$ in $S[\delta]$}
                \State{$P[\delta]$ =  concat($A(F(\tilde{\mathbf{x}}))$)}
    	\EndFor
        \State{\# Dimension of $P[\delta]$ is $M$. Notice $M_{SC}$, $M_{QA}$, $M_{NER}$ in three tasks are different}
	\EndFor
	\State {Return $P_{M\times N}$ for each model $F$}
	\end{algorithmic}
\end{algorithm}
% \vspace{-.15in}

\myparagraph{Extracting Logit Features.} 
For every trigger candidate $\delta \in \Delta$, we insert it to a clean sample set (8 clean samples) with 2 different locations (front location and rear location)\footnote{In NER task, there are three types of attacks. One of the attack 'local', will only be activated if the trigger is in the first half, or the last half of the sentences. So we inject the trigger candidates to front or rear location in order to fully activate the attack.}. 
This creates 16 perturbed samples ($S[\delta]$) per candidate. These samples are processed by the model to gather logits, which are then assembled into a logit feature set for analysis. The feature dimensions vary by task:
% So every trigger candidate $\delta$ constructs to a set of perturbed samples $S[\delta]$, containing ($2 \times 8 = 16$) samples. 
% We infer the perturbed samples $S[\delta]$ with a suspicious model $F$. We select specific model output logits and concatenate them to form the logit features $P[\delta]$. We compute these logits to provide a comprehensive of the model's performance and predictions corresponding to trigger candidates.
In SC task, we select logits from ground truth label and non-ground truth label respectively, which yields to the dimension of logit features $P[\delta]$: $M_{sc}=32$ ($16 \times 2$). 
In QA task, we compute 6 logits related to the start point and the end point of the answer\footnote{Please refer to Appendix \ref{appendix:details_logits} for more details.}, which yields to a feature dimension $M_{qa}=96$ ($16 \times 6$). 
In NER task, we select the logits of all valid tokens in 16 samples, which yields to a feature dimension $M_{ner}=228$ (Notice that the number of valid tokens in 16 samples may be different).

\subsubsection{Justification: Logit Features Reveal Backdoors} \label{sec:Justification_logits}

In this subsection, we validate the efficacy of logit features in distinguishing between clean and backdoored models for various NLP tasks. We start with using true triggers. Furthermore, we show that given a large trigger candidate set $\Delta$, the abnormal logits behavior still exists.

\begin{figure}[ht]
    \centering
    % \vspace{-.25in}
    \includegraphics[width=\linewidth]{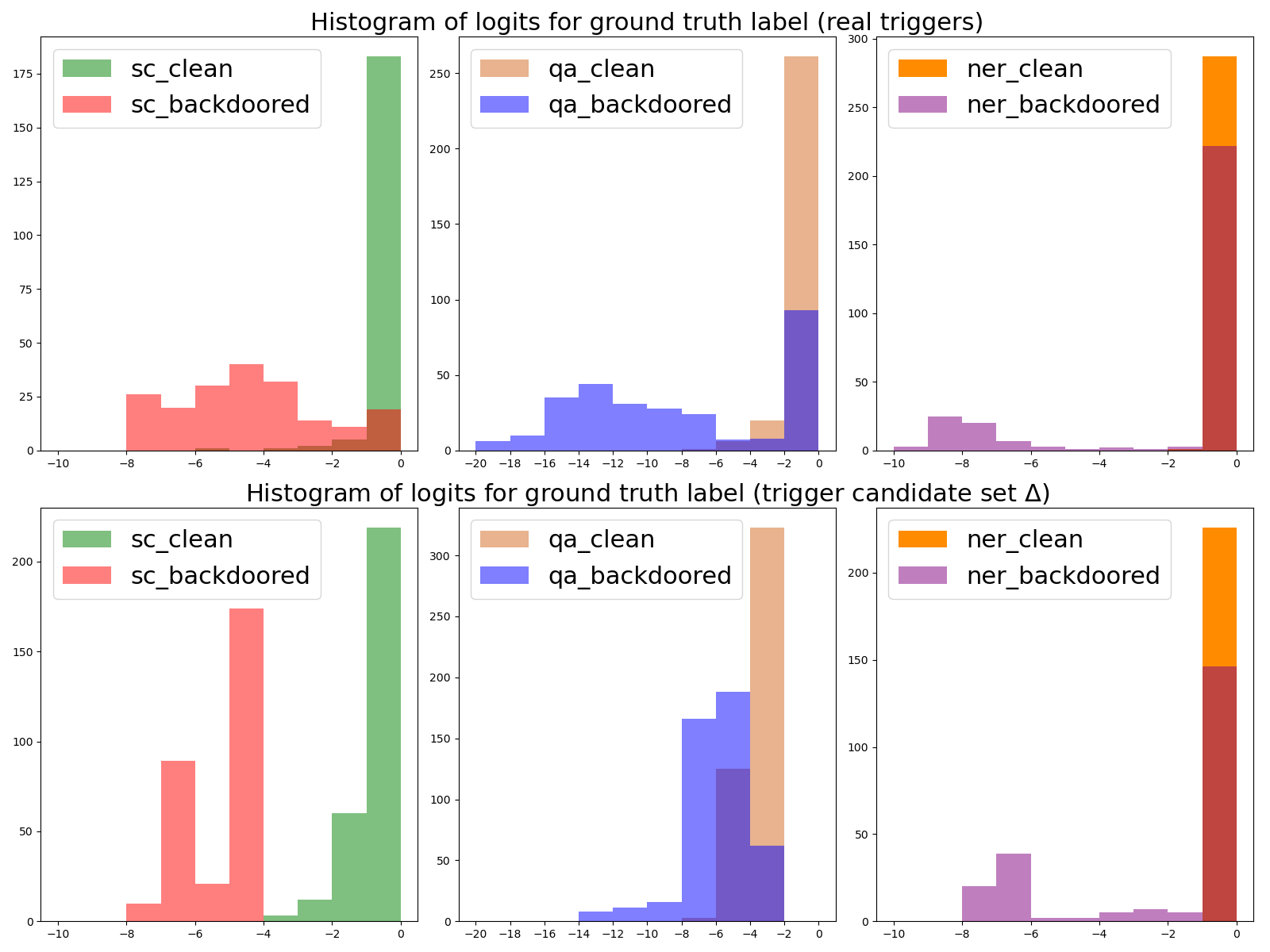} 
    \vspace{-.15in}
    \caption{The histogram illustrates logit distributions for the ground truth label across three NLP tasks, differentiating between clean and backdoored models. x axis is the logit values, y axis is the count of logits in corresponding bins. \textbf{Top Row} shows clear separation in logit values when real triggers are used. \textbf{Bottom Row}, with a large set of trigger candidates $\Delta$ (only display the lowest 0.01\% values), reveals persisting abnormal logit behaviors in backdoored models, demonstrating the robustness of logits as indicators of model integrity.}
    \label{fig:obs1_logits_abnormality}
    % \vspace{-.15in}
\end{figure}

First, we illustrate that given the real trigger, the final layer logits can effectively differentiate clean and backdoored models regardless of the NLP tasks.
We insert the real trigger into aforementioned 16 samples (fixed samples for fixed tasks), and record the logit features (the final layer logits after log-softmax) associated with the ground truth labels (see Figure \ref{fig:logsoftmax} for illustration). As shown in Figure \ref{fig:obs1_logits_abnormality} top row, there are clear differences in logit features between the clean models and backdoored models. This discrepancy is particularly pronounced with the ground truth labels, where backdoored models exhibit significantly reduced logits. This is desired for any successfully backdoored models as they are trained to have such a behavior. This property should commonly hold regardless of the NLP tasks. 
This phenomenon motivates us to use logit features as the potential features for backdoor detection.

Second, we establish that even without exact triggers, the presence of a diverse trigger candidate set $\Delta$ can still elicit abnormal logit responses indicative of a backdoored model. For every trigger candidate $\delta \in \Delta$, we can form $M$ dimension features. For better visualization, we pick the logits of real labels for each sentence. For example, in SC, the sentence 'I like the food.' is a positive sentence, so we picked the logits of positive label. We only plot the lowest $0.01\%$ values due to a large number of features for 62599 trigger candidates. Figure \ref{fig:obs1_logits_abnormality} bottom row shows that the distinct logit distributions for clean and backdoored models are evident, even in the absence of the actual trigger.

 However, the variability in logit dimensions across different NLP tasks and the inherent noise in the logit signals, as illustrated in Figure \ref{fig:a0_intuition}(2) and Figure \ref{fig:analysis3_pooling}(top row), present challenges in developing a unified backdoor detector.
To overcome this and retain the detection power, we introduce a \textit{Representation Refinement} component, which we discuss in the following section. This component is designed to harmonize the logit signals for effective backdoor detection across varied NLP tasks.
%%%%%%%%%%%%%%%%%%%%%%%%%%%%%%%%%%%%%%%%%%%%%%%%%%%%%%%%%%%%%%%%%%%%%%%%%%%%%%%%%%%%%%%%%%%%%%%%%%%%%%%%%%%%%%%%

\begin{figure}[ht]
    \centering
    % \vspace{-.25in}
    \includegraphics[width=\linewidth]{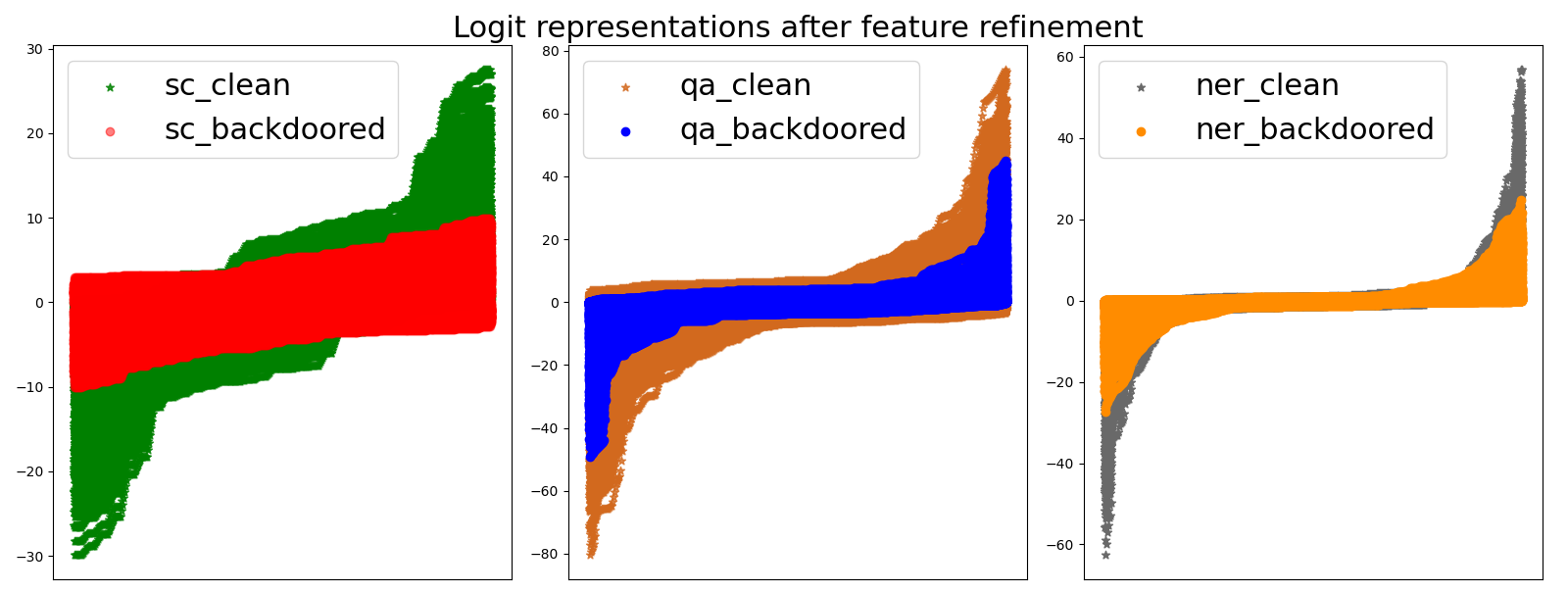} 
    \vspace{-.15in}
    \caption{The refined feature representations effectively differentiate between clean and backdoored models across various NLP tasks. Each color on the figure corresponds to a unique model, with the plotted points indicating individual feature values after refinement in one model. The x-axis labels the feature indices, and the y-axis their corresponding values. The distributions are not only efficient in separation but also exhibit consistency across various NLP tasks, highlighting the effectiveness of the feature refinement process.}
    \label{fig:logits_distribution_before_after}
    % \vspace{-.15in}
\end{figure}

\subsection{Representation Refinement} \label{sec:fr}

In the second component, we refine the logit features into high-quality representations, ensuring consistency across varying architectures and tasks. This critical process enhances the raw logits, facilitating the development of a robust, task-agnostic backdoor detection framework.

The major challenge lies in aligning the logit features from models for different tasks. The logit features from different tasks have varying dimensions. It is very hard to find correspondence; a logit output for SC is not comparable with a logit output for NER. The key insight is that it is indeed sufficient to compare the logit features at a distribution level. This inspires us to propose strategies like qantile pooling and histogram descriptors. 
The quantile pooling technique strategically reduces feature space dimensionality by focusing on its quantiles. The histogram computing further refines this by aggregating logit features into a concise, histogram-based format. These two techniques, together, providing a balanced and comprehensive view of the logits' distribution for effective backdoor detection.

\myparagraph{Quantile Pooling.}
We first propose a quantile pooling scheme. 
We effectively reduce the dimensionality of our feature space while preserving the most critical information embedded in the logits. It enhances the efficacy of our pooling strategy in differentiating between clean and backdoored models.
The quantile index generation is followed by

{\small
\begin{align*}
q^1 &= \left[q_0, q_1, \ldots, q_{\frac{n}{2}-1}\right], \\
q^1_i &= \left(1 + \frac{10}{\frac{n}{2} - 1}\right)^{-i}, \forall i \in \left\{0, 1, \ldots,  \frac{n}{2} - 1\right\} \\
q^2 &= \text{reverse}\left(q^1\right), \\
q &= \left[ \frac{q^2}{2}, \frac{1 - q^1}{2} + 0.5 \right]
\end{align*}
}

\begin{itemize}
    \item \textbf{Non-linear Scale $q^1$:} The formula $\left(1 + \frac{10}{n/2 - 1}\right)^{-i}$ creates a non-linear scale. This allows the indices to be more densely packed at the ends of the distribution and sparser in the middle. This non-linear scale is beneficial when the distribution of logits is not uniform, emphasizing the tails of the distribution where extreme values are present.
    \item \textbf{Balancing the Distribution:} Creating $q^2$ as a reversed version of $q^1$ and then concatenating $\frac{q2}{2}$ with $\frac{1 - q^1}{2} + 0.5$ balances the distribution of indices. The division by 2 and the addition of 0.5 ensure that the indices are evenly distributed across the entire range of logits.
\end{itemize}

The aim is to obtain a set of indices representative of the entire distribution of logits. The generated quantile index ensures that the selected indices capture the essence of the entire distribution. The mathematical expressions are chosen to create a balanced and non-linear distribution of indices, ensuring both common and rare values in the logits are represented. The code implementation can be found in Appendix \ref{appendix:pooling_equation}.
% This approach is useful in scenarios where understanding the entire range of logits distribution is crucial.

% We apply the quantile pooling twice. The first time pooling aims to reduce candidate dimension $N$ dimension to $n/2$ dimension. For a second time, we apply pooling to reduce the logits dimension from $M$ to $m$. 

% In practice, we apply this technique to select 300 indices from the logits data. By doing so, we effectively reduce the dimensionality of our feature space while preserving the most critical information embedded in the logits. This quantile pooling not only aids in simplifying the data structure but also enhances the efficacy of our pooling strategy in differentiating between clean and backdoored models. The selected indices represent a distilled version of the logits, encapsulating the most pivotal aspects needed for accurate model assessment.
% The quantile pooling is instrumental in distilling and concentrating the essence of the logits data, thereby retaining the most informative aspects while reducing dimensionality. 

\myparagraph{Histogram Computing.} 
For our second refinement strategy, we employ histogram binning to analyze the distribution of representations. Each column of length $N$ is sorted and binned into $n/2$ segments, counting the quantity within each. This process yields a dimensionally reduced matrix of size $M \times n/2$, where each column represents a histogram of counts per bin. These histograms uniformly partition the range of each original column, providing a different perspective on the representation distribution. n in our algorithm is a hyper-parameter that specifies the reduced dimension.

\subsubsection{Rationale: Representation Refinement Strategy} \label{sec:poolingstrategy}

\begin{figure}[!t]
    \centering
    % \vspace{-.25in}
    \includegraphics[width=\linewidth]{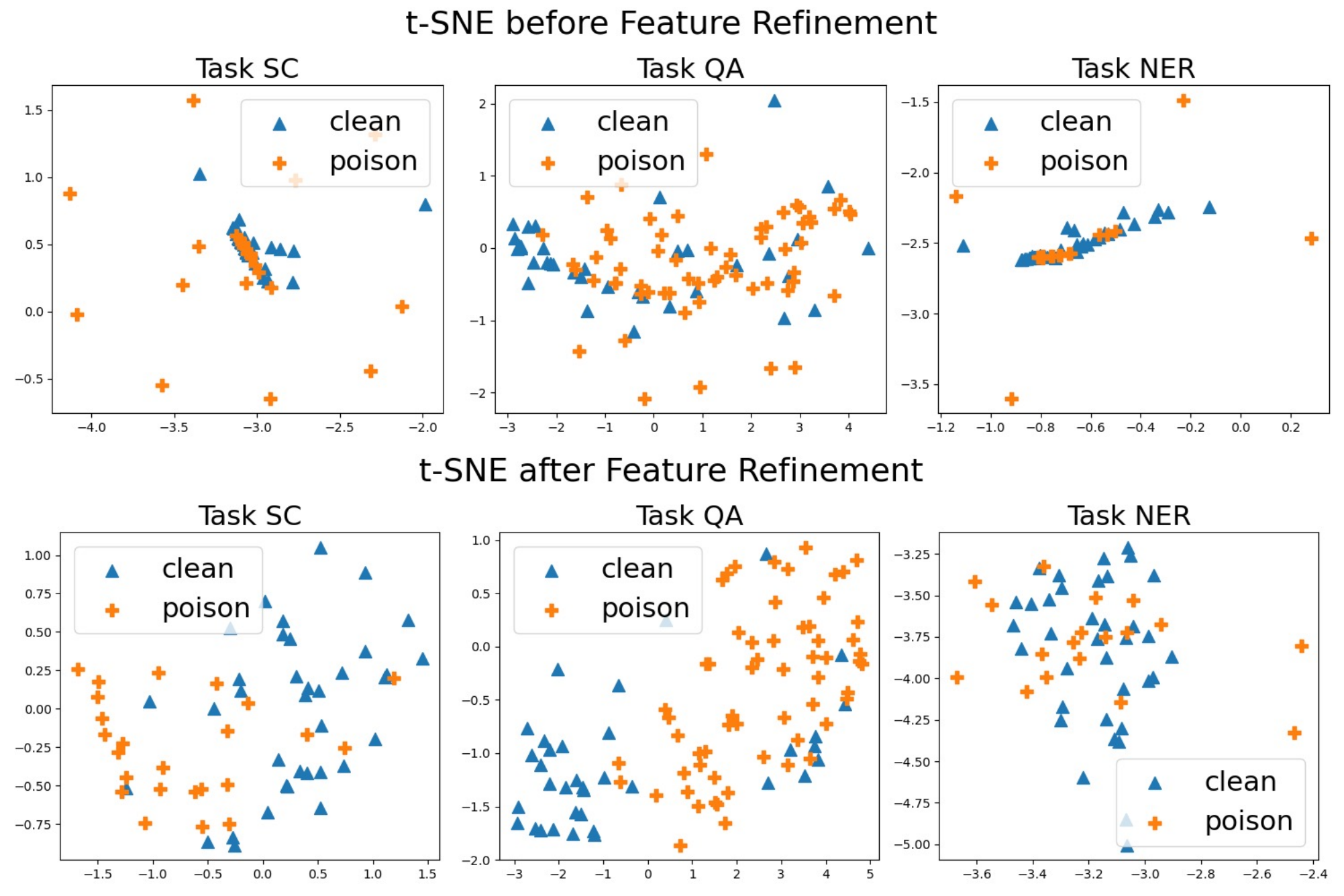} 
    \vspace{-.15in}
    \caption{t-SNE visualization on logit representation before (Top Row) and after (Bottom Row) representation refinement. Each dot indicates one model. By refinement, the representation quality significantly improves.}
    \label{fig:analysis3_pooling}
    % \vspace{-.15in}
\end{figure}

In Figure \ref{fig:logits_distribution_before_after}, we display the distribution of logit representations post-refinement, showcasing their strong discriminatory potential even without further learning.
Complementing this, t-SNE \citep{liu2016visualizing} visualizations in Figure~\ref{fig:analysis3_pooling}(botttom) depict each model's refined logit representation as a distinct point. These visualizations clearly illustrate the heightened separation and enhanced clarity of the refined representations compared to their initial, coarse counterparts. These observations underscore the efficacy of our refinement methods and point towards the feasibility of a backdoor detection algorithm that utilizes these refined representations for training classifiers.
% we plot the representation distribution after the feature refinement. 
% We also visualize these features with t-SNE \citep{liu2016visualizing}. In Figure~\ref{fig:analysis3_pooling}(lower), each point represents the logit representation of a model. We can see that these representations already convey strong discriminative information without any learning procedure involved. Furthermore, the refined representations are much more separated than the coarse ones. These observation not only demonstrates the effectiveness of our refinement methods but also suggest a potential backdoor detection algorithm that trains a classifier using the refined representations.  

%% scripts/obs3/obs3_trigger_set.py

\begin{algorithm}[!h] 
	\caption{Representation Refinement}
	\label{alg:pooling}
	\begin{algorithmic}[1]
		\State {\bfseries Input:} Logit features $P_{M\times N}$, N is the trigger candidate number in $\Delta$, $M$ is the feature dimension, which is various in different tasks
		\State {\bfseries Output:} A unified feature $FR_{m\times n}$, where $m, n$ are identical across tasks
		% \State {Let the candidate set $S=\emptyset$}
		\State{\# Dimension reduction along $N$ dimension}
            \State{$A_{M\times n/2}$ = Histogram($P_{M\times N}$)}
            \State{$B_{M\times n/2}$ = Quantile($P_{M\times N}$)}
            \State{$C_{M\times n}$ = combining $A_{M\times n/2}$ and $B_{M\times n/2}$}
		\State{\# Dimension reduction along $M$ dimension}
            \State{$FR_{m\times n}$ = Quantile($C_{M\times n}$)}
	\State {return refined feature $FR_{m\times n}$}
	\end{algorithmic}
\end{algorithm}
% \vspace{-.15in}

\subsection{Backdoor Detector} \label{sec:td}
After the representation refinement component, we generalize the representation into identical dimension. We then train a Trojan detector, \ie, a MLP classifier, to discriminate whether the suspicious model is a clean model or backdoored model.

\section{Experiments} \label{sec:experiments}

\subsection{Experimental Settings} \label{sec:experimental_settings}
\myparagraph{Datasets and Models.}
% \wl{Statistical table: arch, dataset, task. Other baseline can only do the architecture specific. Our method can ignore the architecture, as long as }
We focus on three NLP tasks: sentence classification task (SC), question answering task (QA) and named entity recognition task (NER). And the model architectures are Roberta \citep{liu2019roberta}, DistilBERT \citep{sanh2019distilbert} and ELECTRA \citep{clark2020electra}, mixed in three tasks. 
We leverage 420 models from the training and test sets of TrojAI NLP-Summary Challenge \cite{trojai:2023, competition_description}. It provides a training set of 210 models, in which 102 are infected with backdoors, and a test set of 210 models , in which 101 are infected with backdoors. The statistics information is shown in Table \ref{tab:stat1}. 
% there are 60 SC models, 96 QA models and 54 NER models in the training set, while 68 SC models, 96 QA models and 46 NER models in the test set. 
The SC models are trained with IMDB dataset \citep{maas2011learning}, the QA models are trained with SQuAD v2 dataset \citep{squad_v2, huggingface_squadv2} and the NER models are trained with CoNLL-2003 dataset \citep{connl_2003}, respectively.
We only consider the standard insertion-based textual backdoor attacks, AddSent \citep{dai2019backdoor} and BadNL \citep{chen2021badnl}, in our experiments. The triggers are words, phrases or sentences. 
A detailed description can be found in Appendix \ref{appendix:exp_details}.

\begin{table}[ht]
% \vspace{-.1in}
\caption{Training and test models statistics.}
\label{tab:stat1}
\begin{center}
\vspace{-.1in}

\resizebox{\columnwidth}{!}{ %% resize table to fix the file

\begin{tabular}{l|ccc|ccc}
\hline
\multirow{2}{*}{} & \multicolumn{3}{c|}{Training} & \multicolumn{3}{c}{Test}    \\
                  & Positive  & Negative  & Total & Positive & Negative & Total \\ \hline
SC                & 24        & 36        & 60    & 31       & 37       & 68    \\
QA                & 60        & 36        & 96    & 54       & 42       & 96    \\
NER               & 18        & 36        & 54    & 16       & 30       & 46    \\ \hline
\end{tabular}

}
\end{center}

% \vspace{-.15in}
\end{table}

\myparagraph{Detection Baselines.}
We implement three textual detection baselines\footnote{Notice that detection and defense are two different research categories, so we do not involve defense baselines here.}, \eg, T-Miner, AttenTD and PICCOLO.
T-Miner \citep{azizi2021t} trains a sequence-to-sequence generator and finds outliers in an internal representation space to identify backdoors. AttenTD \citep{lyu2022study} detects whether the model is a benign or backdoored model by checking the attention abnormality given a set of neural words. PICCOLO \citep{liu2022piccolo} leverages a word discriminativity analysis to distinguish backdoors. 
% For T-Miner and AttenTD, they are designed for sentence classification task, so we only implement them on SC; PICCOLO is designed for SC and NER tasks, so we only implement PICCOLO on SC and NER. 

% \myparagraph{Evaluation Metrics.} We use AUC as our evaluation metrics.

\myparagraph{Implementation Details.}
When training the backdoor classifier, we involve the hyperparameter tuning in order to get a more robust classifier. Hyperparameters include the hidden dimensions number, layers number in each MLP, the quantile pooling interval, Adam optimizer learning rate. We use HyperOPT\footnote{\url{https://github.com/hyperopt/hyperopt}} hyperparameter optimization tool, via 8-fold cross validation on the training set. 
% When reporting results using 8-fold cross validation, the fold for testing was used for deciding the optimal number of optimization epochs, as well as a temperature scaling on the classification logits for confidence calibration for optimal cross entropy. 
% In practice, however, we find that 8-fold crossval performance well.
% As a result, crossval performance may be inflated due to overfitting of hyperparameters. In practice however, we find that 8-fold crossval performance still correlates well with results on the evaluation server.

\subsection{Detection Results}

\myparagraph{Baseline Detection Performance.}
We provide the detection evaluation with existing textual baselines. In their original experiments, T-Miner \citep{azizi2021t}\footnote{Due to the vocabulary size limitation, we only implement T-Miner on the ELECTRA architecture, with totally 19 models.} and AttenTD \citep{lyu2022study} only experiment on SC task, and PICCOLO \citep{liu2022piccolo} experiments on SC and NER tasks. 
% None of the existing detection method designs backdoor detector for the QA task. 
We follow their default experiment settings.
Table \ref{tab:baselines} shows that our TABDet outperforms three baselines in all three tasks. The T-Miner is mainly designed for LSTM-based language models, thus does not perform good on complicated transformer architectures. 
AttenTD's focus on attention abnormalities falls short due to noise and computational inefficiency.
% AttenTD mainly leverage the attention abnormality in backdoored models, however this attention-related feature may be buried among messy noise. Besides, extracting the attention feature is super low in computing efficiency. 
PICCOLO, while performing well on SC and NER, does not leverage other tasks information and lags in detection capabilities. 

% Baseline methods, vs. Ours methods (0.85~9x AUC). 
\begin{table}[ht]
% \vspace{-.1in}
\caption{Detection performance (AUC) compared to baselines. `$-$' indicates not applicable.}
\label{tab:baselines}
\begin{center}
\vspace{-.1in}

\resizebox{0.7\columnwidth}{!}{ %% resize table to fix the file

\begin{tabular}{c|ccc}
\hline
                 & \textbf{SC}   & \textbf{QA}   & \textbf{NER}  \\ \hline
\textbf{T-Miner} & 0.50          & -             & -             \\
\textbf{AttenTD} & 0.60          & -             & -             \\
\textbf{PICCOLO} & 0.87          & -             & 0.72          \\
\textbf{TABDet (Single)} & 0.92        & 0.92        & 0.85         \\ 
\textbf{TABDet}   & \textbf{0.98} & \textbf{0.93} & \textbf{0.86} \\ \hline
\end{tabular}

}
\end{center}
% \vspace{-.15in}
\end{table}

\myparagraph{TABDet Detection Performance.}
% We train TABDet with all models from three tasks, and get a unified detector. Then we test this detector on three tasks. Table \ref{tab:baselines} shows that our TABDet outperforms all the baselines.
TABDet, trained across three NLP tasks, establishes a unified detection approach. As demonstrated in Table \ref{tab:baselines}, it surpasses baseline methods in all tasks.
The performance on NER task is not as good as the performances on other two tasks. That is because the challenge of variability and ambiguity in natural language is particularly prominent in NER. Entities can have different meanings based on their usage and context, and they can easily change once a random trigger candidate is inserted. That makes the backdoor detection on NER task difficulty.

\myparagraph{TABDet Detection in Individual Tasks.}
We also evaluate our framework only with single task. In this setting, we train three individual backdoor detectors for three different tasks. 
In Table \ref{tab:baselines}, Row \textit{TABDet (Single)}: Our TABDet, when applied to single tasks, shows good detection performance, comparing to the performance with other textual detection baselines. This validates the potency of our feature refinement strategy even within the constraints of individual tasks.
However, when compared to the multi-task model training (Table \ref{tab:baselines}, Row \textit{TABDet}), the single-task detectors exhibit slightly reduced efficacy. This highlights the advantage of a multi-task perspective, where TABDet harnesses commonalities across tasks to enhance detection capabilities, as evidenced by the superior performance in multi-task settings.

\subsection{Ablation Study} \label{main:ablation_study}
In this section, we investigate the impact of trigger candidate set size, different pooling strategies, histogram features, and partial trigger effect.

\myparagraph{Impact of Trigger Candidate Set Size. }
% We validate our TABDet with different Trigger Candidate Set $\Delta$. In Table \ref{tab:impact_trigger_set_size}, \textit{2gram}, and \textit{5gram} indicate the trigger candidates in $\Delta$ are two-word and five-word, respectively. With total trigger candidates number 24267 and 62599 candidates, respectively. These two trigger candidate sets are from Google Books Ngram Corpus \citep{michel2011quantitative, lin2012syntactic}. Table \ref{tab:impact_trigger_set_size} illustrates the detection performance boosts while the size of $\Delta$ increases. The overall AUC achieves 0.94 with 5gram, with AUC in individual task 0.98, 0.93 and 0.86 for SC, QA and NER respectively. 
We validate our TABDet with different Trigger Candidate Set $\Delta$. Employing 2gram and 5gram sets from Google Books Ngram Corpus \citep{michel2011quantitative, lin2012syntactic}, with 24,267 and 62,599 candidates respectively, we observed improved detection performance with the increase in $\Delta$ size. In Table \ref{tab:impact_trigger_set_size}, the overall AUC achieves 0.94 with 5gram, with AUC in individual task 0.98, 0.93 and 0.86 for SC, QA and NER respectively. 

% cross-task
\begin{table}[ht]
% \vspace{-.1in}
\caption{Impact of different Trigger Candidate Set $\Delta$.}
\label{tab:impact_trigger_set_size}
\begin{center}
\vspace{-.1in}

\resizebox{0.9\columnwidth}{!}{ %% resize table to fix the file

\begin{tabular}{c|c|cccc}
\hline
\textbf{Trigger Candidate Set}  & \textbf{Number of Triggers}       & \textbf{SC} & \textbf{QA} & \textbf{NER} & \textbf{Overall} \\ \hline
% \textbf{1gram}                  &  23028                            & 0.84        & 0.89        & 0.82         & 0.86             \\
\textbf{2gram}                  &  24267                            & 0.78        & 0.88        & 0.73         & 0.81             \\
\textbf{5gram}                  &  62599                            & 0.98        & 0.93        & 0.86         & 0.94             \\ \hline
\end{tabular}

}
\end{center}
% \vspace{-.15in}
\end{table}

\myparagraph{Impact of Pooling Strategies and Histogram Features.}
First, we examined the effects of different pooling strategies on dimension reduction, contrasting quantile pooling with max, min, and average pooling, as they are common operations in practice. We set the output dimension the same as our quantile pooling. Our findings, outlined in Table~\ref{tab:pooling_strategy_performance}, reveal quantile pooling's superior ability to retain outlier features indicative of backdoors, thereby enhancing detection performance over the other methods. Max/min/average pooling strategies tend to smooth out critical features, diluting backdoor signals, whereas quantile pooling preserves them. Secondly, relying solely on histogram features does not match the efficacy achieved by TABDet's comprehensive approach.

% Table~\ref{tab:pooling_strategy_performance} shows that the quantile pooling strategy significantly improves the detection performance comparing to other three pooling strategies. 
% Max/min/average pooling strategies return max/min/average values, however these values lose most of the backdoor-related information since these pooling strategies smooth the features. When pass the perturbed samples to suspicious models, we would expect only a very small amount of perturbed samples can `activate' the backdoor, with a very low proportion of abnormal yet valuable features. We hope to keep those outlier-style features. But the max/min/ave operations obviously smooth the outliers. 
% On the other hand, if we only keep the histogram features, the results is not as good as TABDet.

% quantile pooling operation returns values at fixed-index quantiles, and it also borrows the histogram features as luxury information, which generalize max/min/average operations. 

% Baseline methods, vs. Ours methods (0.85~9x AUC). 
\begin{table}[ht]
% \vspace{-.1in}
\caption{Ablation study on different pooling strategies and histogram features.}
\label{tab:pooling_strategy_performance}
\begin{center}
\vspace{-.1in}

\resizebox{0.8\columnwidth}{!}{ %% resize table to fix the file

\begin{tabular}{cc|cccc}
\hline
                                                       & \textbf{}    & \textbf{SC} & \textbf{QA} & \textbf{NER} & \textbf{Overall} \\ \hline
\multicolumn{1}{c|}{\multirow{3}{*}{\textbf{Pooling}}} & \textbf{Max} & 0.30        & 0.58        & 0.62         & 0.61             \\
\multicolumn{1}{c|}{}                                  & \textbf{Min} & 0.40        & 0.38        & 0.74         & 0.56             \\
\multicolumn{1}{c|}{}                                  & \textbf{Ave} & 0.49        & 0.38        & 0.63         & 0.59             \\ \hline
\multicolumn{2}{c|}{\textbf{Only Histogram}}                          & 0.73        & 0.78        & 0.82         & 0.78             \\ \hline
\multicolumn{2}{c|}{\textbf{TABDet}}                                  & 0.98        & 0.93        & 0.86         & 0.94             \\ \hline
\end{tabular}

}
\end{center}
% \vspace{-.15in}
\end{table}

\myparagraph{Impact of Partial Triggers.} 
In this ablation study, we explored how partial triggers—snippets of a complete trigger phrase or sentence—can still effectively activate backdoors in models. We found that even two-word from longer triggers can prompt the model to produce the targeted predictions, altering the logit representations significantly. This was empirically validated across three NLP tasks. The robust impact of these partial triggers supports the effectiveness of using a broad and extensive trigger candidate set for backdoor detection, as indicated by our results in Table~\ref{tab:partial_triggers}.

% In this ablation study, we verify the how the partial triggers activate the backdoor. In practice, attackers will use sentences or phrases as triggers to keep semantic integrity of the backdoored inputs and impose significant challenges to detection algorithms that tries to reverse-engineering the ground truth triggers. However, we observe that a model backdoored with such triggers can be activated even if we only input a partial snippet of the trigger sentence. Usually two words from a long sentence or phrase trigger  are sufficient to make the model give targeted prediction and generate significant changes in model's logit representation. 

% We  empirically verify this phenomenon on three NLP tasks (SC, QA, NER). We investigate successfully backdoored models trained using triggers with text length larger than 2.  We inject poisoned sentences to each of these models with a 2-word snippet from the original trigger inserted. From Table~\ref{tab:partial_triggers} we can see these 2-word partial triggers still preserve strong effect in manipulating victim models' outputs. 
% This finding provide us clues that a diverse and large trigger candidate set is reasonable and efficient for backdoor detection. 

% impact of partial triggers
\begin{table}[!htb]
% \vspace{-.1in}
\caption{Attack Performance with Partial Triggers. We report the source label accuracy for SC and NER, report exact match sore for QA.}
\label{tab:partial_triggers}
\begin{center}
\vspace{-.1in}
\resizebox{\columnwidth}{!}{ %% resize table to fix the file

\begin{tabular}{c|c|ccc}
\hline
                                        &                                         & \textbf{SC} & \textbf{NER} & \textbf{QA} \\ \hline
\textbf{Clean Models}                   & \textbf{CleanSamples}                   & 0.98        & 0.92         & 88.75       \\ \hline
\multirow{3}{*}{\textbf{backdoor Models}} & \textbf{CleanSamples}                   & 0.97        & 1            & 88.58       \\
                                        & \textbf{PoisonedSamples-RealTrigger}    & 0.02        & 0            & 19.75       \\
                                        & \textbf{PoisonedSamples-PartialTrigger} & 0.2         & 0.18         & 23.67       \\ \hline
\end{tabular}

}
\end{center}
\vspace{-.15in}
\end{table}

%%%%%%% rebuttal added
\myparagraph{Detection Effectiveness on Advanced Insertion-based Attacks.} 
We also extend our experiments to include two advanced insertion-based textual backdoor attacks, such as EP \cite{yang2021careful} and RIPPLEs \cite{kurita2020weight}\footnote{We implement the backdoor attack with OpenBackdoor toolkit: \url{https://github.com/thunlp/OpenBackdoor}.}. EP and RIPPLES modify different levels of weights/embeddings, such as input word embedding. Given that EP and RIPPLES are primarily designed for sentence classification tasks, we limited their implementation to this specific task, thus this ablation study can only partially validate the detection effectiveness of our TABDet. Details in Appendix \ref{appendix:advanced_ep_ripples}.

Table \ref{tab:advanced_attacks} presents the detection performance of TABDet across different textual backdoor attacks. Our findings indicate that the detection effectiveness of TABDet is comparable across the additional textual backdoor attack baselines. This consistency in performance highlights the robustness of TABDet, attributable to our detection mechanism that focuses on the output logits abnormalities of the models. Irrespective of the textual attack's type, a successfully backdoored model tends to show comparable patterns in the logits of the last layer, specifically in terms of switching the correct label to an incorrect one.

\begin{table}[!htb]
% \vspace{-.1in}
\caption{Detection effectiveness compared with basic attacks (AddSent/BadNL) and advanced attacks (EP/RIPPLES).}
\label{tab:advanced_attacks}
\begin{center}
\vspace{-.1in}
\small
% \resizebox{\columnwidth}{!}{ %% resize table to fix the file

\begin{tabular}{c|ccccc}
                       & \textbf{TP} & \textbf{FP} & \textbf{FN} & \textbf{TN} & \textbf{AUC} \\ \hline
\textbf{AddSent/BadNL} & 10          & 0           & 1           & 9           & 0.95         \\
\textbf{EP/RIPPLES}    & 10          & 0           & 1           & 9           & 0.95        
\end{tabular}

% }
\end{center}
\vspace{-.15in}
\end{table}

\section{Conclusion}
In this paper, we pioneered TABDet (\textit{\underline{T}ask-\underline{A}gnostic \underline{B}ackdoor \underline{Det}ector}), the first unified detector of its kind that operates effectively across three key NLP tasks (sentence classification, question answering, and named entity recognition).
The proposed TABDet utilizes the model's final laye logits, and a unique feature refinement strategy, resulting in a versatile and high-quality representation applicable to sentence classification, question answering, and named entity recognition tasks.
While existing detectors mainly focus on SC and NER tasks, TABDet can detect backdoors from all SC, QA and NER tasks, achieving the new state-of-the-art performance on backdoor detection.

% In this work, we pioneered TABDet, the first detector of its kind that operates effectively across three key NLP tasks. Our method utilizes final layer output logits and a unique refinement strategy, 

% TABDet not only bridges the gap where existing methods falter but also sets a new benchmark for backdoor detection in NLP models.

% We propose TABDet (\textit{\underline{T}ask-\underline{A}gnostic \underline{B}ackdoor \underline{Det}ector}) based on the model's final layer output logits, and propose a novel feature refinement strategy to get a high-quality and common representation across NLP tasks. 
% While existing detectors can only deal with limited tasks, TABDet can detect backdoors from all SC, QA and NER tasks, achieving the new state-of-the-art performance on backdoor detection.

\section*{Limitations}
There are several limitations of our proposed methods. 
1) TABDet is only effective against standard insertion-based attack, and can not deal with more advanced textual backdoor attack such as style transfer based attack \citep{qi2021hidden, qi2021mind}. As future work, we should investigate detection against a broader range of textual backdoor attacks.
2) We only test three popular NLP tasks, namely sentence classification, question answering and named entity recognition tasks, and future work should explore backdoor detection on more NLP tasks.
3) Detection on NER task performs not as good as SC and QA. A more efficient strategy towards NER task should be developed.

\section*{Ethics Statement} \label{sectioln:appendix:ethics}
In this paper, we propose a detection strategy against textual backdoor attacks. Our codes and datasets will be publicly available. We conduct such detection framework only for research purpose and do not intend to harm the community.

\section*{Acknowledgements}
We thank anonymous reviewers for their constructive feedback. This effort was partially supported by the Intelligence Advanced Research Projects Agency
(IARPA) under the Contract W911NF20C0038. The content of this paper does not necessarily reflect the position or the policy of the Government,
and no official endorsement should be inferred.

\bibliography{custom}

\appendix

\section{Appendix}
\label{sec:appendix}

\subsection{Implementation Details in Section \ref{sec:Justification_logits}} \label{appendix:details_logits}
For how to get the logits and plot the Figure \ref{fig:obs1_logits_abnormality}(Top Row), we split into three steps: 1) generate poison samples, 2) use the model do the inference, and record the final layer output logits, 3) format all logits. 

\myparagraph{Step1.} We generate poisoned samples by inserting the real trigger to eight fixed clean samples with two different locations (locations (5, 25)). For clean models, we only use the same eight clean samples without any trigger insertion. In this way, we generate 16 ($2\times 8$) poisoned samples for backdoored models, and 8 samples for clean models.

\myparagraph{Step2.} For backdooreds models, we forward 16 samples to the model and record the final layer out logits. For clean models, we forward 8 samples to the model and record the final layer out logits. We use $log-softmax(logits)$ as logits values. We process logits with \textit{log-softmax} \citep{log_softmax_pytorch} instead of \textit{softmax} \citep{softmax_pytorch} is because the numerical stability and computation efficiency (see Figure \ref{fig:logsoftmax} for illustration). 
For sentence classification (SC) task, we record the logits of the ground truth labels (see Figure \ref{fig:logsoftmax} for illustration). We record one logits for each sample. 
For named entity recognition (NER), since it is classification for tokens, we record the logits of ground truth labels from only valid tokens (labels that are not 0), ignoring useless tokens (0 label). The number of logits depends on how many valid tokens in the samples.
For question answering (QA), we record the logits from start position\footnote{For QA task, since we are using the BERT architecture, and the answer is selected from input text by encoders. So it is classification model, instead of generative model with decoders.}. We record one start position logits for each sample. More specifically, the six logits are: the model's confidence in ground truth start position being the start of the answer, the model's confidence in the ground truth end position being the end of the answer, the model's confidence in the first token being the start of the answer, the model's confidence in the first token being the end of the answer, the model's prediction confidence at the very beginning of the input sequence, the average of previous logits. Basically, we want to incorporate more information through these logits.

\myparagraph{Step3.} For each model, we flatten the aforementioned features into vector. We use all the clean models' features and all the backdoored models' features to plot the distribution in Figure \ref{fig:obs1_logits_abnormality}(top row).

\subsection{Experiments Details in Section \ref{sec:experimental_settings}} \label{appendix:exp_details}

\myparagraph{Dataset and Models Description.}
Our experiments leverage models from TrojAI NLP-Summary Challenge \citep{trojai:2023}, the detailed dataset and models description can be find \citet{competition_description}.
There are 420 models in the original test set, and we only select the first 210 test set in our experiment setting. In this way, we have 210 models in training set, and 210 models in test set, with same dataset size.

\myparagraph{Attack Configurations.}
In TrojAI NLP-Summary Challenge \citep{trojai:2023}, there are several attack configurations. For the textual backdoor attacks across three NLP tasks, there are totally 17 trigger configurations: 
1) 10 types triggers for QA: `context\_normal\_empty', `context\_normal\_trigger', `context\_spatial\_empty',
`context\_spatial\_trigger', `question\_normal\_empty', `question\_spatial\_empty',
`both\_normal\_empty', `both\_normal\_trigger', `both\_spatial\_empty', `both\_spatial\_trigger',
2) 3 types triggers for NER: `global', `local', `spatial\_global', and 
3) 4 types triggers for SC: `normal', `spatial', `class', `spatial\_class'.

For backdoor attacks against NER tasks, we only select trigger type `global' and `spatial\_global', removing `local' trigger type. The `local' trigger means that the trigger is inserted directly to the left of a randomly selected label that matches the trigger source class, modifying that single instance into the trigger target class label. In this specific and advanced `local' attack, it's hard to `activate' the backdoor pattern. Our study mainly focus on the insertion-based backdoor attacks, and `local' trigger type does not belong to the insertion-based attack, so we remove this specific type during testing.

\myparagraph{Hyperparameter Tuning.}
For both types of pooling, hyperparameters including the hidden dimensions and number of layers of each MLP, the quantile pooling interval, Adam optimizer learning rate and number of epochs can be automatically determined through hyperparameter search.

\myparagraph{A Broad Scope of Related Work.}
Although the field of security research encompasses a broad array of topics \cite{liu2024please, liu2023riatig, liu2023slowlidar, wang2022adversarial, chen2023dark, zhang2023byzantine, li2024comprehensive, liang2023model, liang2021omnilytics, zhuang2022defending, chen2023ride}, this study narrows its focus to the exploration of backdoor learning (detection). 
Compared to the evolution of neural networks in various domains \citep{wang2020topogan, wang2021topotxr, lyu2022multimodal, lyu2019cuny, pang2019transfer, dong2023integrated, wu2023fineehr, wu2023bottrinet, wu2023botshape, wang2022computationally, wang2023dual, chen2023hypergraf, li2023bubble, chen2022relax, chen2022explain, zhang2021distractor, srivastava2023instance, huang2023mental, zhan2022deepmtl, wu2023enhanced, qian2024next, zhuang2024understanding, zhuang2022robust, xie2022deepvs, xie2024scaling, liu2023financial, zhou2023towards, gupta2022learning}, our research primarily focuses on textual transformer-based architectures, which have become predominant in most NLP applications.

\subsection{Implementation Details of Detection Effectiveness on Advanced Insertion-based Attacks} \label{appendix:advanced_ep_ripples}

In Section \ref{main:ablation_study}, part `Detection Effectiveness on Advanced Insertion-based Attacks', we also extend our experiments to include more sophisticated insertion-based textual backdoor attacks, such as EP \cite{yang2021careful} and RIPPLEs \cite{kurita2020weight}. We introduce the details of this ablation study. Given that EP and RIPPLES are primarily designed for sentence classification tasks, we limited their implementation to this specific task.

We trained 10 backdoored models, and 10 clean models, with the SST-2 dataset. To maintain consistent experimental conditions, we also generated 10 backdoored models using the AddSent and BadNL attack methods, as mentioned in our original manuscript, keeping all other settings identical.

\subsection{Google Books Ngram Corpus}

Google Books Ngram Corpus \citep{michel2011quantitative, lin2012syntactic}. It is build by a sequence of n-grams occurring at least 40 times in the corpus, and this corpus contains $4\%$ of all books ever published in the world. The n-grams covers the space of English text efficiently, which would provide a strong inductive bias for finding backdoor triggers that are English words. We use 5-gram trigger candidate set for all three tasks.

\subsection{Use Log-softmax over Softmax}

% Explain why we use log-softmax instead of softmax/logit
Unlike the bounded softmax output, log-softmax lies in the range of $(-\infty, 0)$ and numerically benefit the computation (see Figure \ref{fig:logsoftmax} for illustration).  Furthermore, the log-softmax representation gives a non-positive score for each input sentence. The smaller the score, the more likely it triggers the backdoor behavior. A classifier trained on log-softmax representations can better identify backdoor model's output.

\subsection{Quantile Pooling Operation} \label{appendix:pooling_equation}

% \begin{lstlisting}

We use the following equation to decide our index selection when we implement the quantile pooling strategy, as described in Section \ref{sec:fr}. We show the code implementation of quantile pooling as follows:

\begin{verbatim}
q=
((1+10/(N//2-1))**(-torch.arange(N//2-1)))
    .tolist()+[0] 
    # N//2 length list
q2=q[::-1]
q=torch.Tensor(q)
q2=torch.Tensor(q2)
q=torch.cat((q2/2,(1-q)/2+0.5),dim=0) 
    # lead to a sorted index
\end{verbatim}

% \end{lstlisting}

\subsection{Visualization on Final Feature Representation.}

\begin{figure}[ht]
    \centering
    % \vspace{-.25in}
    \includegraphics[width=\linewidth]{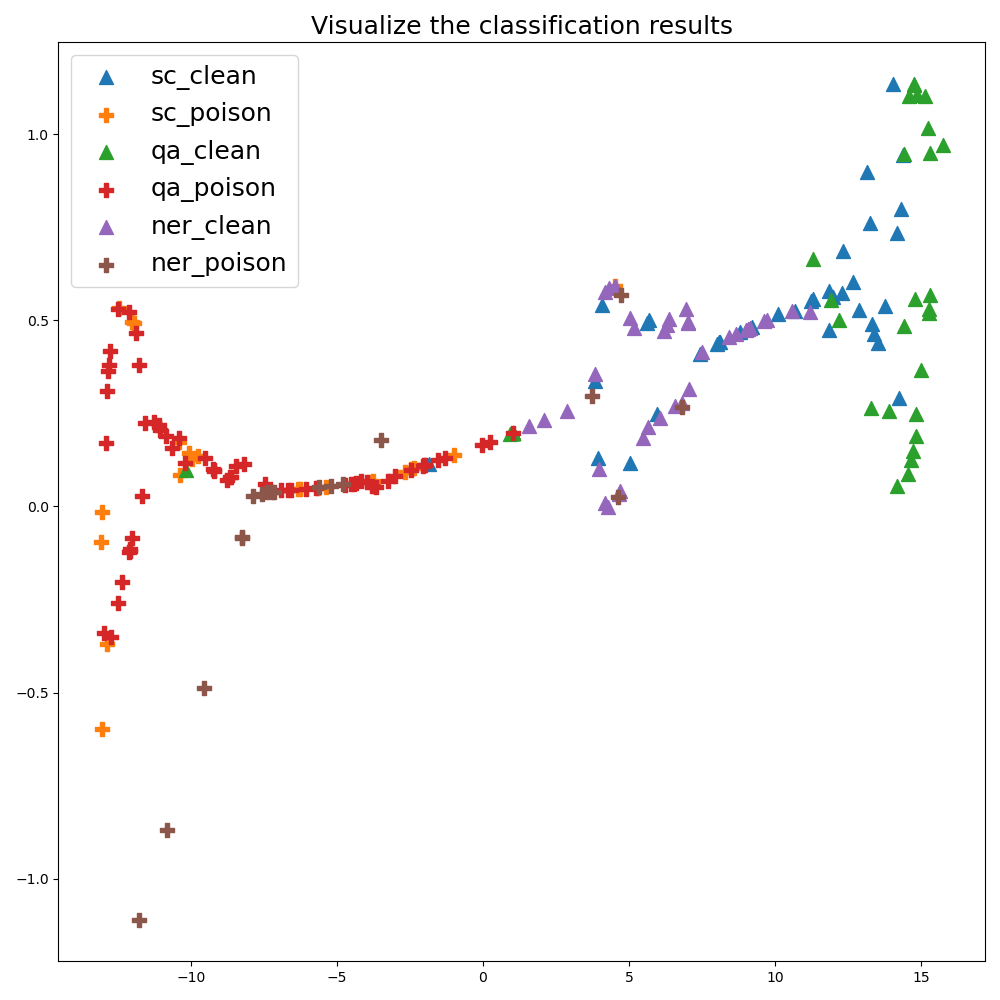} 
    % \vspace{-.25in}
    \caption{Visualization on Final Feature Representation.}
    \label{fig:tsne_final}
    % \vspace{-.15in}
\end{figure}

Fig.~\ref{fig:tsne_final}, t-SNE on backdoor detector's final layer outputs. With our representation refinement strategy, the backdoor detector learns a very good feature representation.

% TSNE final

\end{document}